\documentclass[10pt]{article}

\usepackage[top=2.0cm, bottom=1.5cm, inner=2.0cm, outer=2.0cm]{geometry}
\usepackage{float}
\restylefloat{table}
\usepackage{enumerate}
\usepackage{color,graphicx,colortbl}
\usepackage{paralist,parskip}
\usepackage{booktabs, multirow, makecell}
\usepackage{subfigure, bm, comment}
\usepackage{amsmath,amsfonts,amssymb}
\usepackage{mathabx,fixmath,algorithm, amsthm} 
\usepackage{pstool}
\usepackage{psfrag}
\usepackage{hyperref}
\usepackage{siunitx}
\usepackage{cite}
\usepackage{algpseudocode} 
\usepackage{bm} 
\usepackage[table,xcdraw]{xcolor} 
\usepackage{tabularx}
\usepackage{mathtools}

\usepackage{natbib}
\usepackage{csquotes}
\usepackage{cleveref}

\newcommand{\N}{\mathbb{N}}

\newcommand*{\vv}[1]{\mathbf{#1}}

\newtheoremstyle{spaced}
{8pt}
{2pt}
{\itshape}
{}
{\bfseries}
{.}
{2pt}
{}

\theoremstyle{spaced}

\definecolor{Gray}{gray}{0.9}

\long\def\hide#1{}
\newcommand{\rev}[1]{{{\color{black}#1}}}
\newcommand{\rrev}[1]{{{\color{black}#1}}}
\newcommand{\rrrev}[1]{{{\color{black}#1}}}

\title{\rrev{On model selection for scalable time series forecasting in transport networks}}

\author{Julien Monteil, Anton Dekusar, Claudio Gambella, Yassine Lassoued, Martin Mevissen}
\date{}

\begin{document}
	
	\pagenumbering{arabic}
	\pagestyle{plain}

	\maketitle

	\begin{abstract} 
		The transport literature is dense regarding short-term traffic predictions, up to the scale of 1 hour, yet less dense for long-term traffic predictions. The transport literature is also sparse when it comes to city-scale traffic predictions, mainly because of low data availability. \rrev{In this work, we report an effort to investigate whether deep learning models can be useful for the long-term large-scale traffic prediction task, while focusing on the scalability of the models}. We investigate a city-scale traffic dataset with 14 weeks of speed observations collected every 15 minutes over 1098 segments in the hypercenter of Los Angeles, California. \rrev{We look at a variety of state-of-the-art machine learning and deep learning predictors for link-based predictions, and investigate how such predictors can scale up to larger areas with clustering, and graph convolutional approaches.}  \rrev{We discuss that modelling temporal and spatial features into deep learning predictors can be helpful for long-term predictions, while simpler, not deep learning-based predictors,} achieve very satisfactory performance for link-based and short-term forecasting. The trade-off is discussed not only in terms of prediction accuracy vs prediction horizon but also in terms of training time and model sizing. 
	\end{abstract}
	
	
	\section{Introduction}\label{sec:intro}

Traffic prediction in urban transport networks is a central task for the real-time operation of transportation systems, such as route planning, route guidance, on-demand mobility services~\cite{SIMONETTO2019208}. In principle this task can be achieved with the help of an increasing large volume of observed traffic data that can be made available through, e.g.,  on-road sensors, GPS data, cameras, social media~\cite{ZhuSurvey19}. In reality, the access to such data is limited as big traffic data sets are generally owned by specific companies and deemed as proprietary information and a valuable source of business. 

Despite the limited availability of such data sets, there has been a constant interest in the transport and intelligent systems literature to develop forecasting methods for the prediction of traffic conditions, mainly traffic speeds. Traffic prediction is inherently a challenging problem, because of the high non-linearity of traffic variables\rev{, the sparsity of data \cite{hofleitner2012learning, abadi2014traffic, 6878453},} the spatio-temporal dependencies, the short vs long term dynamical effects, and the computational challenges of its application to large-scale networks. In order to be suitable for practical implementation, predictions should be obtained in a reasonably fast and accurate manner, and the models used to make such predictions should scale up with the size of the geographic area under consideration.
Short-term time series traffic forecasting has been shown to be well handled via autoregressive models, in particular the univariate (seasonal) autoregressive integrated moving average (SARIMA) class of models~\cite{box2015time}. Such models are able to capture the strong autocorrelations in the evolution of speeds, as well as the cyclic patterns of traffic over time, and typically outperform pure linear regression, historical average, and non-parametric regression~\cite{smith2002}, for \rev{predictions of no more than 1 hour}. 
This class of models can be extended to the consideration of spatial dependencies with the vector autoregressive (VAR) and vector autoregressive with moving average (VARMA) models and their use may result in an improvement in the accuracy of the predictions \rev{\cite{chandra2009predictions}}.  \rev{Possible improvements to accuracy can be found via multivariate approaches, although they might be computationally demanding~\cite{STATHOPOULOS2003121,ghosh2009multivariate}.}
 Gradient boosting has also been adopted for short-term traffic predictions \cite{ZHANG2015308}, demonstrating to require modest data preprocessing, to fit complex nonlinear relationship, and to provide interpretable results.
The advent of deep learning techniques and their easy API access through the many wrappers to TensorFlow have led researchers into systematically contemplating them as plausible solution to the spatio-temporal traffic forecasting problem.
\cite{Lv2015} proposed a stacked autoencoder model to consider the spatial and temporal correlations of traffic flow inherently.  \rev{Spatio-temporal predictors can be enhanced with attention functions \cite{do2019effective}, or ensembled in convolutional models \cite{liu2020spatio}.}
Short-term traffic flow prediction is performed in \cite{POLSON20171} on a 21-loop network via a multi-layer perception models. Spatio-temporal relations are captured by solving a prediction selection problem. 
The long term dependency of traffic predictions has also been taken into account in the Long Short-Term Memory Neural Network model \cite{MA2015187}, \rev{\cite{vinayakumar2017applying}}.
Integrating spatial and exogenous dependencies of traffic predictions into LSTM architectures is an active research topic \cite{KE2017591}.

In the deep learning applications, particularly speech, video, and image processing, the use of convolutional units was shown to achieve remarkable results in terms of accuracy and scalability, see e.g.~\cite{lecun1995convolutional,ji20123d, oquab2014learning, liang2015recurrent}. Convolutions are typically very successful to consume data with a clear grid-structured topology. In the Residual Neural Network framework of \cite{zhang2017deep}, the convolutional units model the spatial properties of the crowd flows. 
 
Graph Convolutional Neural Networks (GCNN) \cite{BrunaGCCNN} extend the Convolutional Neural Network architectures to cope with spatial features on graphs. A spectral construction of GCNN, which builds upon the properties of convolutions in the Fourier domain is applied to traffic forecasting in \cite{ijcai2018}.
 The model is effectively applied to networks with up $1,026$ stations of speed measurement for predictions in a $45$ minutes horizon.\rev{~\cite{zhang2019wavelet} proposed the use of wavelet decomposition to capture multiple time-frequency properties in a similar GCNN-based spatio-temporal model, and this beat state-of-the art results on a dataset for short term traffic prediction}. \rev{An effort to make the convolutional approach more scalable appears in~\cite{kipf2016semi}, and was applied successfully to traffic prediction in~\cite{zhao2019t}. The GCNN combined with gated recurrent units achieved remarkable results in terms of robustness and accuracy for predictions with up to $1$ hour ahead.} Finally, the GCNN of \cite{ying2018graph}, is of interest because of the large size of the network under test: by applying localized graph convolutions, recommender systems with 7.5 billion training examples are built.  

In this work we \rrev{first benchmark several state-of the art time-series predictors for the long-term prediction task, including contextual average model, autoregressive models, decision tree models, multi-layered perceptrons, and long short-term memory networks. The accuracy of the various predictors is first evaluated at the link level, i.e. small-scale, and the benefits of the deep learning techniques are highlighted as we predict farther in time. Second, we look into a number of possibilities to scale up the deep learning techniques to the network level, and observe that prediction accuracy is often achieved at the expense of computational performance. A spatio-temporal deep learning scalable predictor relying on a graph convolutional neural network architecture is proposed, achieving a satisfactory accuracy/performance trade-off.}

The contributions of our approaches with respect to the current literature are: \rrev{(i) we run a first comparative benchmarking study of state-of-the art traffic prediction models for the long-term large-scale regression task, up to $3$ hours  ahead, with thousands of road network links; (ii) we investigate how deep learning can be used to scale up traffic predictions at the network level and leverage a scalable GCNN architecture to seamlessly incorporate spatio-temporal fetures into the predictor; (iii) while noting that the results are achieved on a specific dataset, we provide recommendations for practitioners about the use of deep learning for traffic prediction: for short term prediction, standard autoregressive models and tree-based models may be used; for long-term prediction, deep learning techniques do provide an accuracy lift, and should be considered; if scalability is a concern, i.e. maintaining 1 model per network vs 1 model per network link, the presented GCNN, and variations of it, should be considered, as it provides an elegant way to capture temporal and spatial dependencies in a unique network model, while achieving the best accuracy for long term prediction.}

	\section{The dataset}\label{sec:exp-frame}

\subsection{The dataset}

The data set was made available to us through The Weather Company, an IBM business. We do not have agreement to share the data, however we believe that the findings presented in this work are valuable to the research community, as traffic data sets become increasingly available (see, e.g., the repository \cite{trafficdata}). 
For the purpose of this work we selected a geographic area of Los Angeles inner city using the bounding box defined by the 4 latitude and longitude coordinates: north 34.047252, south 34.000675, east -118.244915, west -118.338785. This bounding box includes 1220 road segments, belonging to local roads, arterial roads, and highways.
We gathered 14 weeks of speed observations data from Sunday 22 July 2018 00:00 to Saturday 28 October 2018 23:45. The speed observations come for all the segments at a frequency of 15 min, but there are some processing delays leading to uneven frequencies. Such speed observations are computed by averaging speeds of the vehicles transiting via a given segment over a window of 15 minutes. Speed observations are reported in km/h throughout the paper. 


\subsection{Processing methodology}

The data set requires some preprocessing prior to its utilization. We should mention that we proceed with the following steps: (i) removal of default average speeds used as instantaneous speeds; (ii) linear interpolation for missing speeds; (iii) time step regularization with linear interpolation to deal with shifted timestamps.


\begin{figure}[htbp!]
\centering
\includegraphics[width=0.49\textwidth]{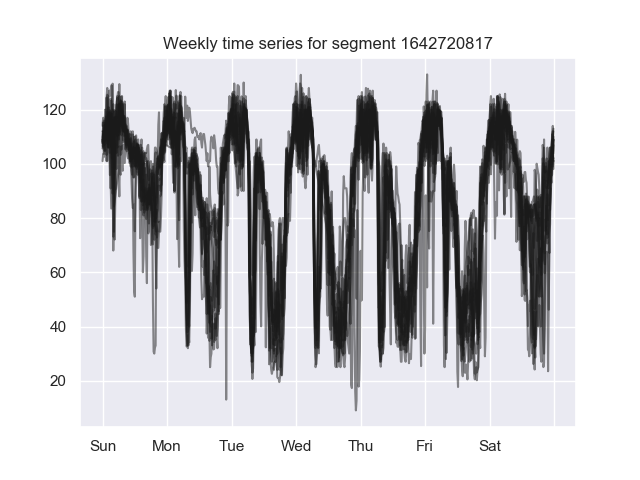}
\caption{Cumulative weekly time series for the 14 weeks of data, segment 1642720817. Speed values are given in km/h.}
\label{fig:3}
\end{figure}
 Note also that, if used, the low-pass filter needs to be carefully chosen, i.e. so as to remove high frequency but not to smooth out congestion transition regimes and therefore not to alter the physics of traffic speeds evolution. 
Figure~\ref{fig:3} shows the cumulative weekly time series of the corrected speed observations. Since human mobility is a highly predictable phenomenon, up to 93\% following the work of~\cite{song2010limits}, it is interesting to look at the weekly patterns observed across the 14 weeks of the data set. The figure clearly shows the existence of predictable patterns in the data set for the same selected segment. We must remark that such patterns are not always that obvious across segments.
Finally, only the segments with missing values that account for less than 20\% of the time series were considered. Starting from 1220 segments in the Los Angeles selected bounding box, this boils down to 1098 segments with enough observations. 

\subsection{Stationarity and periodicity}

\begin{figure}[htbp!]
	\centering
	\includegraphics[width=0.49\textwidth]{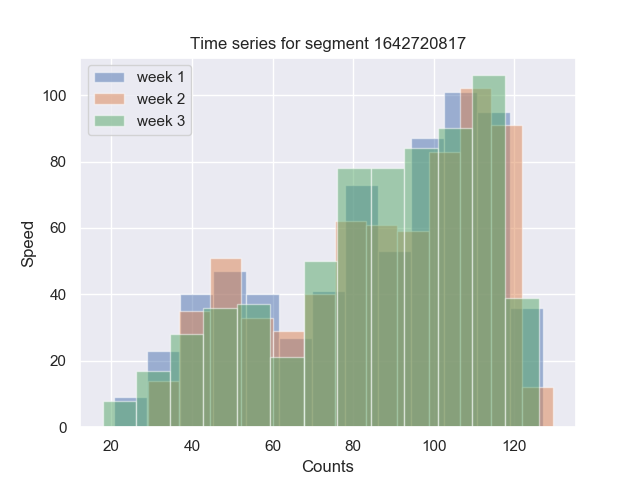}
	\includegraphics[width=0.49\textwidth]{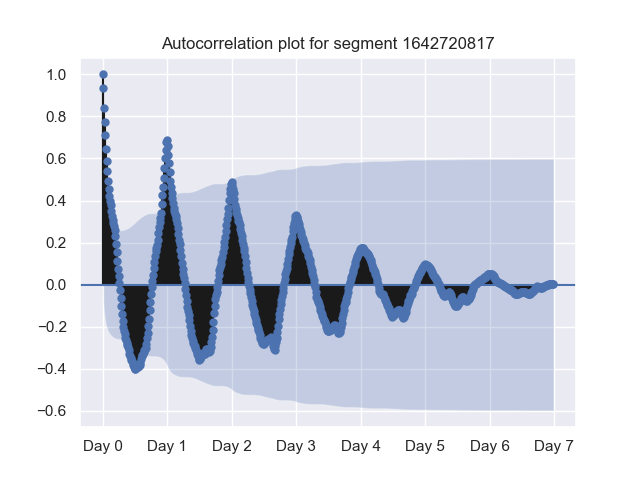}
	\caption{Distributions of speeds for the first 3 weeks of the data set, segment 1642720817 (top). Autocorrelation plot for the first week of the data set, segment 1642720817 (bottom).  Speed values are given in km/h.}
	\label{dataanalysis}
\end{figure}

A first analysis prior to applying autoregressive models consists in looking at the time series observations, particularly at their stationarity and periodicity properties. 
The stationarity aims at saying if a preprocessing step is required before using autoregressive models, e.g. the integrated part of the class of seasonal autoregressive integrated moving average models~\cite{box2015time}. A common way of investigating the stationarity of the time series is to look at speed distributions across weeks. Here, as expected for the speed traffic variable, in Figure~\ref{dataanalysis} (top) we see little variation of such distributions. 
The periodicity can be used for feature selection in feed-forward neural networks, or as a necessary preliminary step to applying seasonal autoregressive integrated moving average models~\cite{box2015time}. The autocorrelation plot Figure \ref{dataanalysis} (bottom) is of great use, and unsurprisingly exhibits strong positive correlations for periods of 24 hours and negative correlations for periods of 12 hours. This preliminary time series analysis is representative of all the segments in the considered bounding box, and will be used in the discussions of the next sections. 

	\section{Time series traffic prediction}\label{sec:pred}

Time series forecasting consists in predicting the most likely measurements in a given time horizon of $1,\ldots,h$ time steps given the previous available $m$ observations. We write $\vv{x}_t\in\mathbb{R}^N$ the vector of measurements of all $N$ links at a given time $t$, $\vv{x}_t=\left[x_{1,t},\ldots,x_{N,t}\right]^T$. The links are connected through the graph of the road network denoted $\mathcal{G}=(\mathcal{N},\mathcal{L},\mathcal{W})$, with $\mathcal{N}$ the set of all nodes, $\mathcal{L}$ the set of all links indexed by $ 1,\ldots,N$.

The univariate forecasting problem for link $i \in \mathcal{L}$ consists in finding the estimates $\hat x_{i,t+1},\ldots ,\hat x_{i,t+h}$ maximising the probability $p(x_{i,t+1},\ldots ,x_{i,t+h}|x_{i,t-m+1},\ldots,x_{i,t})$,, whereas the multivariate forecasting problem searches for the estimates $\vv{\hat x}_{t+1},\ldots ,\vv{\hat x}_{t+h}$ maximising the probability $p(\vv{x}_{t+1},\ldots ,\vv{x}_{t+h}|\vv{x}_{t-m+1},\ldots,\vv{x}_{t})$.

We will see that for each selected algorithm, for obvious scalability reasons, we do not feed all the available $m$ observations to make predictions, hence input feature selection is an important part of the algorithm configuration~\cite{977291}, and we shall rely partly on domain knowledge for this. 

\subsection{Clustering time-series}\label{sec:cluster-t-s}

We explore time-series clustering as a potential prior step to the design of forecasting models. Note that the incorporation of clusters into the predictors will be done in ad-hoc or integrated way depending on the predictor itself. Note we do not perform spatio-temporal clustering despite recent works advocating for its promises~\cite{pascale2015spatiotemporal,lopez2017revealing}, as we aim to perform a regression task and not a classification one. Instead, we explore time series clustering for model sizing, as well as to investigate how it can help refining the prediction accuracy, particularly in the case of spatio-temporal predictors. 

Specifically, for clustering we consider the standardized observations on the validation week for all $1098$ links of the LA area and use a wavelet decomposition technique. We applied the Daubechies 6 wavelet to the time series to obtain the features of each link, which are the relative contributions of maximum resolution level $J$ introduced in \cite{antoniadis2013clustering}. In detail, given the wavelet coefficients $\mathbf{d}_j = \{d_{j, 0}, \dots, d_{j, 2^{j}-1}\}_{j = 0, \dots, J-1}$ of time series $z$, the relative contributions of the scale $j$ to the global energy of $z$ via wavelet decomposition are
$$ \text{rel}_j = \frac{\lVert \mathbf{d_j}\rVert^2}{\sum_{j=0}^{J-1} \lVert \mathbf{d_j}\rVert^2} \quad  j=0, \dots, J-1.$$
The $3$ features with lowest frequency are retained, as \cite{antoniadis2013clustering} shown them to be the most informative ones. Finally, we use a $K$-means clustering with input given by the features selected, in order to determine the clusters of links. The optimal number $K=9$ of clusters is detected by examining the elbow curve and silhouette scores obtained for a number of clusters up to $50$.

\subsection{Baseline time-series predictors}

We now present state-of-the-art predictors and how we used them in the context of this work. Note that the list of baseline predictors we consider here is not meant to be exhaustive, as other univariate models such as tree-based ensemble methods~\cite{ZHANG2015308}, generative additive models~\cite{dominici2002use} could be used. In this section, $t-d$ will denote the time step of the same time $t$ the day before, and $t-w$ the time step of the same time the week before.

\subsubsection{Contextual average model} 

this model is the simplest possible, and consists in computing the plain average over the history of data points. With $N_w$ being the number of weeks for which we have link-based historical data, we have $\forall i\in\mathcal{L}$,
  $$ x_{i,t+k} = \frac{\sum_{j=1}^{N_w} x_{i,t-j\cdot w+k}}{N_w}, \quad k =1, \dots, h.$$

\subsubsection{Autoregressive models}

for the literature on autoregressive models for time series forecasting the interested reader can refer to the book of~\cite{box2015time}. 

In this work we consider the univariate autoregressive integrated moving average ARIMA-$(p,d,q)$ model which is defined as a cascade of two models: 
\begin{align*}
y_{i,t}&=\left(1-L\right)^dx_{i,t}, \\
 \left(1-\sum_{j=1}^p\phi_i L^j\right)y_{i,t}&=\left(1+\sum_{j=1}^q\theta_iL^j\right)\epsilon_{i,t},\end{align*}
with $L$ being the lag operator, i.e. $L^j y_{i,t}=y_{i,t-j}$, and $\epsilon_{i,t}$ are the residuals assumed to be i.i.d. and sampled from a normal distribution with zero mean and standard deviation $\sigma$.  The seasonal variant of the ARIMA model is written SARIMA-$(p,d,q)\times(P,D,Q)_S$, with $S$ the lag of the seasonality, typically of 7 days in the traffic prediction case. We can remark that such models are mostly used for short term forecasting, i.e. 1 step ahead forecasting, as the variance of the forecast error increases with the time horizon~$h$. Note also that a multivariate extension of these models is the vector autoregression VAR-$(p)$ model which describes the evolution of stacked endogeneous variables. As this approach suffers scalability issues and its endogenenous assumption is not valid for the small city regions for which it scales, we leave it aside in this work.

\rev{
\subsubsection{Support Vector Regression}
support-vector machines (SVM) have been introduced for binary classification, but received considerable attention for regression tasks \cite{smola2004tutorial}. 
The goal of support vector regression (SVR) is to find a function $f(x) = w \cdot \mathrm{\Phi}(x) + b$ that can approximate the relationship between features $x_i \in \mathbb{R}^n$ with label $y_i \in \mathbb{R}$, on training points $i=1, \dots, l.$
The non linear transformation $\mathrm{\Phi}$ maps $R^n$ to a higher dimensional space. SVR is achieved by minimizing the regression risk $C\sum_{i=1}^l(\mathrm{\Gamma}(f(x_i) - y_i)) + 1/2\|w\|^2$, with $C$ hyperparameter, and $\mathrm{\Gamma}$ cost function. Kernel functions are typically used to solve the optimization problem. We have followed \cite{wu2004travel} to apply SVR to the traffic speed prediction.}

\subsubsection{Gradient boosting (GB)}
we look into ensemble models and more precisely into boosting algorithms, as a way to get a better performance and get a better bias. In boosting, weak learners are trained on sampled parts of the dataset and combined in an iterative way so as to get a strong learner. In this work, we selected gradient boosting~\cite{friedman2001greedy} where we use $[1,\ldots,M]$ sampled datasets. $\forall i\in[1,\ldots,M]$, we compute the loss function $L^i(y_j,F^{i}(\mathbf x_j))$ typically of the least square form $\sum_{j=1}^n(y_j-F^{i}(\mathbf x_j))^2$, and we fit an estimator $H^i$, typically a shallow regression tree, to the gradient of the loss function $\partial L^i/\partial \mathbf x_j$. Then the predicted estimator becomes $F^m(\mathbf{x})=F^0(\mathbf{x})+\rho\sum_{i=1}^m H^i(\mathbf{x})$, where $F^0(\mathbf{x})$ is an initial guess and $\rho$ is the learning rate. 

\rev{
\subsubsection{Deep Learning time-series predictors}

we work with the windowed Multi-Layer Perceptron (MLP) and the Long Short Term Memory (LSTM) neural networks. Such models have proven to be relevant for time series forecasting, and a consensus is that recurrent networks should be tried if the simpler windowed feed-forward neural networks fails~\cite{gers2002applying}. 

\paragraph{Link-based windowed MLP}

the idea is to feed in the relevant input features only to the feed-forward network that will in turn learn complex dependencies with such features. Let $w_n$ the selected time window for the previous hours, $w_d$ the selected time window for the day before, and $w_w$ the selected time window for the week before. Hence the vector of relevant speed observations is the stacked vector made of the weekly window $X_{w,i,t} = \left[x_{i,t-w-w_w+1},\ldots,x_{i,t-w+w_w}\right]^T$, the daily window $X_{d,i,t} =\left[x_{i,t-d-w_d+1},\ldots,x_{i,t-d+w_d}\right]^T$, and the previous observations window $X_{n,i,t} =\left[x_{i,t-w_n+1},\ldots,x_{i,t}\right]^T$. In our case, we also consider a vector of contextual features $C_t$. In principle we have 2 contextual features, namely time of the day, day of the week, that we feed into the network under a cyclical form. This is done to help the network condition the weights on the context.

Hence, the vector of input features $I_{i,t}$ at a given time $t$ is written $I_{i,t}=[X_{w,t}^T,X_{d,t}^T,X_{n,t}^T,C_t^T]^T$.
This input passes through the network, layer by layer, until it arrives at the outputs of dimension $h$. For instance, with a vector of input features in $\mathbb{R}^{d_0}$, with $d_0=w_n+2w_d+2w_w+2$, the output of the first layer will be $O_{1,i,t}=f_{1,\theta_1}(I_{i,t})=f(W_1\cdot I_t+b_1)$, with $W_1\in\mathbb{R}^{d_0\times d_1}$ and $b_1\in\mathbb{R}^{d_1}$ the weights and biases to be learned, and with $f_1$ being the activation function. The final output $O_{L,i,t}$ is of dimension $h$, with $L$ is the number of layers in the feed-forward network. The training will aim to provide the vector of parameter estimates $\hat\theta_1,\ldots,\hat\theta_L$, and then the learned model can be scored to provide the best estimate, $\forall i\in\mathcal{L}$, $$\left<\hat x_{i,t+1},\ldots ,\hat x_{i,t+h}\right>=f_{L,\hat\theta_L}\circ\ldots\circ f_{1,\hat\theta_1}(I_{i,t}).$$

\paragraph{Link-based LSTM}

although recurrent neural networks are the natural way of modelling time series, they inevitably suffer from the gradient vanishing problem~\cite{hochreiter2001gradient} when it comes to learning long-term dependencies. In practice, this can be avoided with LSTM networks~\cite{hochreiter1997long} which essentially introduces a cell state that keeps relevant information flowing along the network. For the given input feature $I_{i,t}$, an example of forward pass through a LSTM layer is written: 
\begin{align*}
f_t&=\sigma_g(W_f I_{i,t}+U_fh_{t-1}+b_f),\\
i_t&= \sigma_g(W_i I_{i,t}+U_ih_{t-1}+b_i),\\
o_t&=\sigma_g(W_o I_{i,t}+U_oh_{t-1}+b_o),\\
c_t &= f_t * c_{t-1}+i_t* \sigma_h(W_cI_{i,t}+U_ch_{t-1}+b_c),\\
h_t&=o_t* \sigma_h(c_t),
\end{align*}
where $*$ is the Hadamard product, $\sigma_g$ and $\sigma_c$ are the sigmoid and hyperbolic tangent function, $h_t$ is the vector of hidden states that can in turn be fed into a dense layer to obtain the vector of outputs of dimension $h$. Note that in this work, only link-based LTSM models are considered given their already large computational training time. Note also that we do not consider the promising attention-based RNN, \rev{see e.g.~\cite{do2019effective}}.

\subsection{Scalable Deep Learning time-series predictors}
In this section, we propose a few Deep Learning architectures, in an attempt to move away from the link-based models presented so far, towards spatio-temporal models. From a practical perspective, this means fewer models to train and maintain, while there is the hope of a better performance given that we train on spatial dependencies. 
}

\subsubsection{Area-based brute force MLP (B-MLP)}

a naive approach for MLP architectures consists in augmenting the input feature space by the number of links in the network, hence we have, $\left<\vv{\hat x}_{t+1},\ldots ,\vv{\hat x}_{t+h}\right> = f_{L,\hat\theta_L}\circ\ldots\circ f_{1,\hat\theta_1}(I_{t})$, where the stacked input feature vector is now written $I_{t}=\left[I_{1,t}^T,\ldots,I_{N,t}^T\right]^T$. Note that this approach does not increase the number of parameters comparing to the link-based MLP model as the set of weights is applied by the framework to all stacked feature vectors. This is the advantage of this method as the model size stays the same, for equal number of network layers.

\subsubsection{Cluster-based MLP (C-MLP)}

we make use of the time series clustering of Section~\ref{sec:cluster-t-s} by learning a model for each cluster $k$ in the set of clusters $\mathcal{C}$. Denoting $\mathcal{L}_k$ the set of links in the cluster $k$, we treat the relevant speed observations of each link in $\mathcal{L}_k$ as additional data points to learn the unique model for $k$. Then after having estimated $\hat\theta_1,\ldots,\hat\theta_L$ for the given number of layers $L$, then the learned model can be scored in a univariate fashion to provide the best estimate, and we have $\forall i\in\mathcal{L}_k$, $$\left<\hat x_{i,t+1},\ldots ,\hat x_{i,t+h}\right>=f_{L,\hat\theta_L}\circ\ldots\circ f_{1,\hat\theta_1}(I_{i,t}).$$

\subsubsection{Spatio-temporal deep learning predictor}

the brute force MLP and cluster-based MLP offer relatively okay scalability at the expense of accuracy, as for the B-MLP a brute-force average model is in fact learned per link, and as for the C-MLP inter cluster dependencies are not modelled and spatial information, if considered, is embedded into temporal information. In order to circumvent such problems, here we introduce a scalable deep learning predictor that make use of spatio-temporal convolutions in the road network graph.

The architecture we use gets its inspiration from the  hyper large-scale use of GCNN for web-scale recommender systems~\cite{ying2018graph}, and from the use of a MLP-derived CNN architecture for city-wide crowd prediction~\cite{zhang2017deep}. As we are more interested into scalability and accuracy than proper mathematical derivation, we step away from the elegant graph convolutional works that use graph Laplacians to define convolutional operators in the Fourier domain, \rev{e.g. as proposed initially in~\cite{defferrard2016convolutional} as well as} in the context of traffic predictions~\cite{ijcai2018}. \rev{In fact, making use of such a convolutional operator requires computing the eigenvalues and eigenvectors of the Graph Laplacian which is simply not scalable for networks with more than thousands of nodes. The work of~\cite{kipf2016semi} later applied in~\cite{zhao2019t} proposes a first-order approximation, which only requires to know the adjacency and degree matrices of the graph. Their defined operator has 2 main differences with our proposed operator: first, due to the presence of the adjacency matrix in the operator, each layer acts upon its neighbours only, hence with a two-layer model the output of a node can only be affected by nodes in a 2-hop neighborhood, and that means many layers are needed to capture long spatial dependencies; and second, the operator does not capture distances between links. }


In the architecture we present, the input features combine the road network graph structure with the relevant traffic observations, as per the MLP link-based model. For a given link $i$, we consider the $k$ nearest incoming links, $(i^{\text{in}}_1, \cdots, i^{\text{in}}_k)$, and $k$ nearest out-going links, $(i^{\text{out}}_1, \cdots, i^{\text{out}}_k)$, where $k \in \N^*$. \rrrev{This set of neighbouring links is $\mathcal{N}^i$}. 

\begin{figure}[htbp!]
	\begin{center}
		\includegraphics[width=0.50\textwidth]{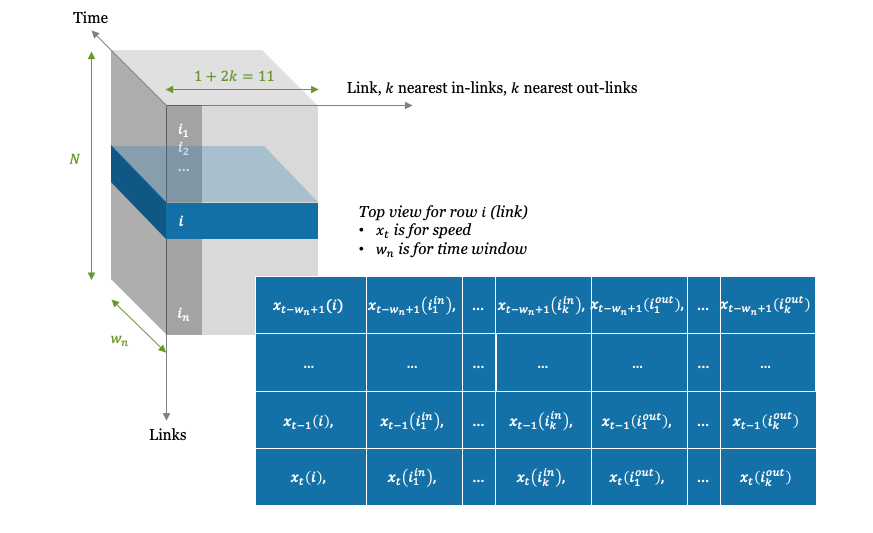}
		\caption{\rrrev{View of the GCNN Input Tensor $T^n$, with detailed link view $T^{n,i}$. The dimension of the input tensor is of $N\times(2k+1)\times w_n$, with $N$ number of links, $k$ is the number of incoming and outgoing links considered, and $w_n$ is the number of previous observations considered.}}
		\label{fig-cnn-input-tensor}
	\end{center}
\end{figure}

\rrrev{Based on the selection of the neighbouring links, at a given time $t$ and for a link $i$, three matrices of inputs features are built from the vectors of relevant speed observations $X_{w,i,t}$, $X_{d,i,t}$, $X_{n,i,t}$. They are written as $T^{w,i}=[X_{w,i,t}^T,X_{w,i^{\text{in}}_1,t}^T,\cdots,X_{w,i^{\text{in}}_k,t}^T,X_{w,i^{\text{out}}_1,t}^T,\cdots,X_{w,i^{\text{out}}_k,t}^T]^T$, $T^{d,i}=[X_{d,i,t}^T,X_{d,i^{\text{in}}_1,t}^T,\cdots,X_{d,i^{\text{in}}_k,t}^T,X_{d,i^{\text{out}}_1,t}^T,\cdots,X_{d,i^{\text{out}}_k,t}^T]^T$, \\ $T^{n,i}=[X_{n,i,t}^T,X_{w,i^{\text{in}}_1,t}^T,\cdots,X_{n,i^{\text{in}}_k,t}^T,X_{n,i^{\text{out}}_1,t}^T,\cdots,X_{n,i^{\text{out}}_k,t}^T]^T$. The input tensors $T^{w}$, $T^d$, $T^n$ are then built by concatenating these matrices along the link axis, as illustrated in Figure~\ref{fig-cnn-input-tensor}. For the sake of clarity, we focus on the input tensor $T^{n,i}$. The dimension of the input tensor are $N\times(2k+1)\times w_n$, with $N$ number of links, $k$ is the number of incoming and outgoing links considered, and $w_n$ is the number of previous speed observations considered. We set $k = 5$, thus considering 10 neighbouring links for the link $i$ in total.}

\rrrev{Given this input tensor, the defined convolution operator is the succession of 2 operations:
\begin{align*}
\label{convop}
S^i&=\text{ReLU}(T^{n,i}\star W_i +b_i), \forall i\in\{1,\cdots,N\},\ \text{(step 1)}\\
\tilde{T}^{n,i}&=\text{Concat}(S^i,S^n),  \forall n\in\mathcal{N}^i, \forall i\in\{1,\cdots,N\}, \ \text{(step 2)}
\end{align*}

with $W_i$ and $b_i$ the kernel filter weights and biases of dimension $(2k+1)\times I$, $*$ the standard kernel convolutions, $I$ is an integer which size can vary across convolutional layers. Hence $S^i$ are vectors of dimension $w_n-I+1$. Concat is the concatenation operation that reconstructs the matrices $\tilde{T}^{n,i} $ along the space axis, hence the updated tensor $\tilde{T}^{n,i}$ are of dimension $(w_n-I+1)\times(2k+1)$. This customized operator enables the propagation in time and space of the speed observations, while at each operation shrinking the dimension along the time axis of the tensor $\tilde{T}^n$.} The weights and bias of the 2D convolutions can be either different across channels (links) or shared across links.
	 
%
%

	Note the weights are learned over the full vector which is composed of the link together its neighbours, differently to~\cite{ying2018graph} where an aggregation step is introduced with a dense layer that is used to combine the representation of the link with the representation of its neighbours, and that we tried this feature without observing significant differences in the test set accuracy. Note also that the first convolutional operations are done separetely on the three input tensors \rrrev{$T^n$, $T^d$, $T^w$}, before being concatenated together, \rrrev{along the time axis. This means that the number of convolutions will be indicated as $x+y$, meaning $x$ separate convolutions, followed by $y$ common convolutions}.  Finally, the tensor is augmented with the contextual features (time of the day, day of the week) and fed into a final dense layer. 
	
	\begin{figure}[htbp!]
		\begin{center}
			\includegraphics[width=0.50\textwidth]{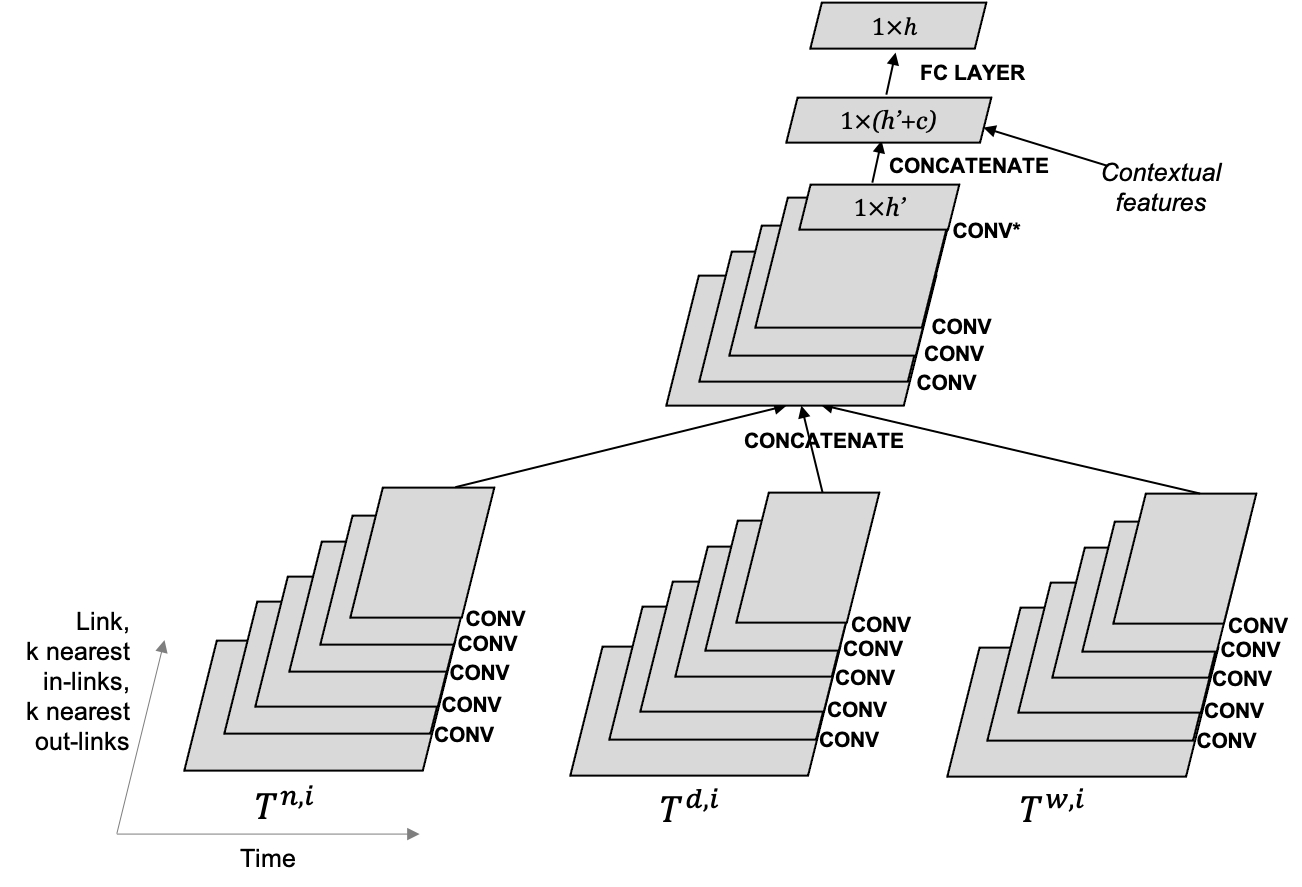}
			\caption{\rrrev{Architecture of the proposed GCNN. It is presented for a given link $i$. \textbf{CONV} designates the defined convolutional operator, and \textbf{CONV*} the defined convolutional operator without the concatenation step (step 2). The convolution operator shrinks the dimension of the matrices (tensor) along the time axis. After stacking the contextual features, a fully connected layer enables to recover a vector of dimension $h$, of the length of the prediction horizon.}}
			\label{fig-cnn-input-archi}
		\end{center}
	\end{figure}
	
	\rrrev{A 5+4 architecture is presented in Figure~\ref{fig-cnn-input-archi}. For the sake of simplicity, the visualization is from a link perspective, noting that the same operations are done for each link in parallel. The inputs tensors that are fed to the GCNN are respectively $T^n$, $T^d$, $T^w$. Five separate convolution operators are being applied. Then the updated tensors are concatenated along the time axis. Four additional convolutional operators are then applied, the last one without the concatenation step (step 2). This leads to a $N\times h'$ matrix. After concatenation of the contextual features of dimension $c$ along the time axis we get $h' + c$ features per link and then a fully connected layer is applied to get to the output matrix of dimension $N\times h$: the regression loss per link can be computed.}


	\section{Computational results}\label{sec:results}

\subsection{Experimental methodology}

\begin{figure*}
	\includegraphics[width=0.99\textwidth]{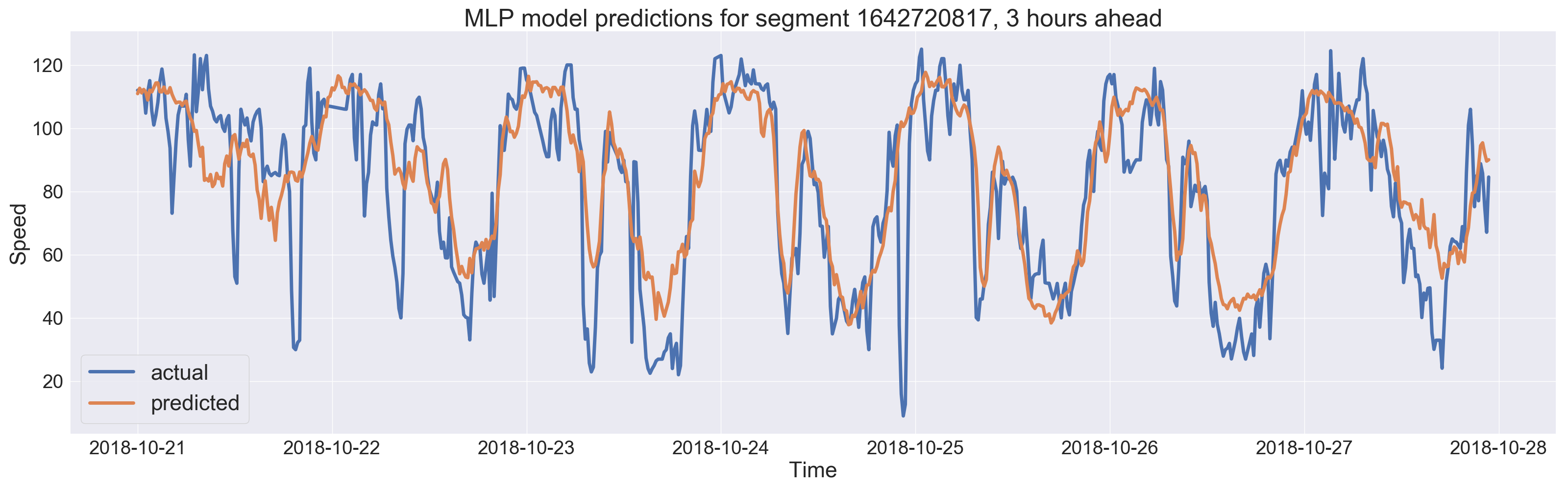}
	\caption{MLP test results, segment 1642720817.  Speed values are given in km/h.}
	\label{fig:6}
\end{figure*}

We have 14 weeks of data from 2018-07-22 00:00 to 2018-10-27 23:45. We propose the following evaluation methodology of the predictors. We split the data in train, validation and test datasets: 12 weeks training until 2018-10-13 23:45, 1 week validating from 2018-10-14 00:00 until 2018-10-20 23:45, and 1 week testing from 2018-10-21 00:00 until 2018-10-27 23:45. The validation dataset is used for hyperparameters tuning of all predictors via grid search.

For the link-based predictors, we present the Root Mean Square Error (RMSE) results over a time horizon of 3 hours ($h=12$) for 50 considered links, randomly sampled from the 1098 links under consideration. Such evaluation is done with the processed dataset as per Section~\ref{sec:exp-frame}.

For the area-based predictors, we present the average of the RSME results over a time horizon of 3 hours ($h=12$) for the previously sampled links. Note that the predictor is trained for the  1098 links but only evaluated on the 50 sampled links, for the sake of clarity of the comparison. Again, such evaluation is done with the processed dataset as per Section~\ref{sec:exp-frame}.

Note that the $\text{RMSE}_h$ is defined as
\begin{equation*}
\rev{\text{RMSE}_h = \sqrt{\frac{\sum_{i=1}^{L}\sum_{t=1}^{N}(\hat x^h_{i,t} -x^h_{i,t})^2}{NL}},}
\end{equation*} 
with $L=50$ being the number of sampled links, and \rev{$N$} being the total number of tested timestamps within the testing week, i.e. with $N=672$. \rev{We compute RMSE for each time step $h$, thus $\hat x_{i,t}$ and $x_{i,t}$ take values for the corresponding $h$.} We remark that the product $NL$ is high and that focusing on such a long testing period will have a smoothing effect on the RMSE results, as some methods compare relatively similarly for long time periods, and as some links are more predictable than others. We choose to display those results as we aim to provide an objective accuracy metrics over a long enough time period (a week), as opposed to a metrics targeting more specific time periods.

We summarize here the hyperpameters determined for each predictor:
\begin{itemize}
	\item For the autoregressive models, the simple AR model with no integrated or moving average term was selected with $28$ previous observations \rev{following the methodology in~\cite{box2015time}}.
	\rev{
		\item SVR implementation is based on the approach described in \cite{wu2004travel} with the window size of 16 and the RBF kernel instead of a linear one.
	}
	\item GB has been implemented in XGBoost \rev{considering techniques from \cite{ZHANG2015308}} 
	, with maximum depth of a tree of $5$, $200$  trees to fit, $16$ previous observations considered in the feature set, and the training has been performed with a learning rate of $0.1$.
	\item For LSTM \rev{(described in \cite{hochreiter1997long})} $24$ previous observations have been considered in the feature set, and an architecture with $2$ layers and $192$ hidden states has been trained with mini-batches of size $50$, learning rate $0.002$ and $40$ epochs.
	\item For the link-based MLP predictor, \rev{(inspired by \cite{zhang2017deep})} the feature set contains observations of the previous $6$ hours ($w_n=24$ observations in a sample), of a time window of $+/-2$ hours around the same time the day before ($w_n=8$, hence 16 observations), and of a time window of $+/-1$ hour around the same time the week before ($w_w=4$, hence 8 observations). The number of layers has been set to $5$, and the training is conducted with batch size of $150$, learning rate of $0.0005$, and weight decay factor of $0.0002$. The number of epochs is optimized for each link via early stopping, and is consistently lower than $200$. For the brute-force approach, the number of layers was increased to $10$.
	\item For the cluster-based MLP, the feature set contains $w_n=16$, $w_d=8$, $w_w=4$. The training has been done on $80$ epochs and a batch size of $100$ for all clusters, on the same MLP architecture of the link-based predictors.
	\item For the GCNN, due to the higher computational time, we use the same feature set as the link-based MLP algorithm, with $w_n = 24$, $w_d=8$ and $w_w=4$. \rrev{The number of convolutional layers has been determined to be $5+4$, meaning 5 convolution operations were separate across the feature spaces $X_{n,t}$, $X_{d,t}$, and $X_{\omega,t}$ and 4 convolution operations common after stacking the vectors together. The first convolution is done with as many filters as links, and the rest of the convolutions with only 1 filter.} The Kernel parameter values $I$ were chosen to slowly decrease from 5 to 2 with strides equal to 1.  The training was done on $70$ epochs with a batch size of $150$. 
	\item \rrev{The contextual features are fed as time of the day, day of the week. The addition of the weather conditions was tried but did not lead to better test accuracy. We strongly think that this is due to the flat weather conditions observed in Los Angeles for the duration of the experiment. We also tried to incorporate the road type categorical feature in the models, either as additional contextual feature or as a way to select the number of filters in the GCNN, however that did not improve the test accuracy. We do recommend to always try including such features when available.}
\end{itemize} 

As an example, on Figure~\ref{fig:6} we show predictions produced by the MLP model for the chosen segment on the test data set. This figure represents traffic speed predictions made at a time of three hours ahead of the recent observations. Obviously, the MLP model was able to capture weekly patterns and provide reasonable predictions on a long term perspective.

\subsection{Trade-off: prediction accuracy vs prediction horizon}

Table~\ref{tab:nrmselink1} summarizes the results, and the lowest RMSE reported among all predictors is marked in bold. The baseline indicates the contextual average model. All predictors are more accurate than the baseline in the first $4$ time steps. GB and MLP are consistently better than the baseline, especially in the first $3$ time steps, where GB achieves an RMSE reduction of more than $10\%$, and MLP of more than $20\%.$ AR reports a $2.15\%$ increase of RMSE in step $5$, and the deviation to the baseline increases for longer term predictions, reaching a $14.90\%$ increase at the last time steps.  
\rev{Despite SVR achieves results comparable to the best of the neural networks on the first step, the overall growth of RMSE is still high and at the end of the time horizon the results are comparable to the baseline while at the last two time steps RMSE values are higher than the baseline (less than $1\%$).}
B-MLP and C-MLP are more accurate than baseline for short-term predictions, while they report a greater RMSE after the $6$-th time step.
C-MLP is consistently more accurate than AR in all time steps, apart from the first one, where C-MLP is $2.14\%$ less accurate. The difference between the RMSE of C-MLP and MLP reaches maximum value of $17.21\%$ on the last time step. Interestingly the GCNN model reports the lowest RMSE for predictions on $2$ hours ahead and the following ones. LSTM provides the best predictions for step $7$ (along with GCNN), and performs slightly worse than GCNN for the longer term prediction. This shows that long-term predictions are more accurate when spatio-temporal dependencies are considered by GCNN. Similarly, LSTM perform better than link-based and cluster-based spatial predictors for long-term predictions, and our interpretation is that it can capture more complex temporal dependencies caused by e.g. spatial dependencies.


\begin{table}
	\centering
	\caption{$\text{RMSE}_h$ reported for each prediction method and time step $h=1, \dots, 12$, on the raw dataset. \rev{The baseline predictors reported are: SVR \citep{wu2004travel}, GB \cite{ZHANG2015308}, AR \cite{box2015time}, MLP (inspired by \cite{zhang2017deep}), LSTM \cite{hochreiter1997long}. }}
	\label{tab:nrmselink1}
	\resizebox{0.49\textwidth}{!}{\begin{tabular}{c|cccccccccccccccc}
			%
				%
			\multirow{2}{*}{\rev{Method}} & \multicolumn{12}{c}{\rev{$\text{RMSE}_h$}} \\
			\cline{2-13}
			& \rev{$1$} & \rev{$2$} & \rev{$3$} & \rev{$4$} & \rev{$5$} & \rev{$6$} & \rev{$7$} & \rev{$8$} & \rev{$9$} & \rev{$10$} & \rev{$11$} & \rev{$12$}\\
			\midrule
			\rowcolor{Gray}
			Baseline & 8.37 & 8.38 & 8.38 & 8.38 & 8.38 & 8.38 & 8.38 & 8.38 & 8.38 & 8.38 & 8.38 & 8.39\\
			\midrule	
			\rowcolor{Gray}
			AR & 5.59 & 7.15 & 7.82 & 8.26 & 8.56 & 8.78 & 8.96 & 9.14 & 9.29 & 9.42 & 9.54 & 9.64\\
			\midrule	
			\rowcolor{Gray}
			\rev{SVR} & \rev{5.60} & \rev{7.05} & \rev{7.61} & \rev{7.91} & \rev{8.06} & \rev{8.15} & \rev{8.21} & \rev{8.28} & \rev{8.33} & \rev{8.36} & \rev{8.39} & \rev{8.41}\\
			\midrule
			\rowcolor{Gray}
			GB & \rev{\textbf{5.42}} & \rev{6.85} & \rev{7.39} & \rev{7.61} & \rev{7.76} & \rev{7.83} & \rev{7.90} & \rev{7.92} & \rev{7.93}  & \rev{7.96} & \rev{7.98} & \rev{8.00}\\
			\midrule
			\rowcolor{Gray}
			MLP & 5.58 &\textbf{6.82}&\textbf{7.27}&\textbf{7.50}&\textbf{7.63} &\textbf{7.71} & 7.79 & 7.86 & 7.91 & 7.97 & 8.03 & 8.08\\
			\midrule
			\rowcolor{Gray}
			LSTM & 5.60 & 6.87 & 7.31 & 7.53 & 7.67 & 7.73 & \textbf{7.77} & 7.82 & 7.85 & 7.86 & 7.88 & 7.90\\
			\midrule
			\rowcolor{Gray}
			B-MLP & 6.86  & 7.45  & 7.77  & 8.02  & 8.25  & 8.45  & 8.66  & 8.87  & 9.08  & 9.26  & 9.43  & 9.57 \\
			\midrule
			\rowcolor{Gray}
			C-MLP &    5.71  & 7.09  & 7.62  & 7.96  & 8.24  & 8.45  & 8.66  & 8.86  & 9.04  & 9.21  & 9.35  & 9.47 \\
			\midrule
			\rowcolor{Gray}
			GCNN & 5.72 & 7.08 & 7.41 & 7.57 & 7.68 & 7.74 & \textbf{7.77} & \textbf{7.80} & \textbf{7.82} & \textbf{7.85} & \textbf{7.87} & \textbf{7.87}
	\end{tabular}}
\end{table}         

\subsection{Trade-offs: training time, model sizing, recommendations}

This Section looks at the operational cost of selecting a given model e.g. in a production system. Table~\ref{tab:pred-stats} summarizes training times, model sizes and number of links covered the models. We remind that $N$ is the total number of links in the network, $h$ the length of the prediction window. Hyperparametrization is not included in the training times, but we should keep in mind that the hyperparametrization effort grows with the number of models. Training times and model sizes are specified per model while the models spans different number of links. The AR model provides the best training time and smallest model size, however without surprise the prediction accuracy becomes worse than the baseline after 4 steps only. 
\rev{The SVR model training time is still very short, compared to the neural network-based models. The drawback of SVR is the number of models needed (1 model per link and per prediction horizon), together with the large storage space needed. The GB model training time is 1.8 larger comparing to SVR but still quite short. The model suffers from the same drawbacks as SVR but to a less extent} The GB model produces the best results on the very first step. The LSTM model training time is more than 6 times greater than the MLP one, and the model size is 47 larger than MLP. This is a high computational cost to pay, however the LSTM provides low RMSEs for long-term predictions. We experimented the spatial models for scalability reasons (1 or $C$ models for the whole network), however B-MLP, C-MLP, are heavy in training and are not able to produce better accuracy results than the other predictors, excepted for the GCNN model. Their advantage lies in the decreased number of models, hence reduced hyperparametrization effort.

From a practical perspective, the MLP model may be the best trade-off considering training time, model size and overall prediction accuracy in the prediction horizon of $3$ hours. Figure~\ref{fig:6} shows that the accuracy at 3 hours, for a representative link, is still satisfactory despite other models providing slightly better accuracy. But a single MLP model still covers only one link which may be an inconvenience to maintain, when the road network grows to 1M+ links (which corresponds to the Greater Los Angeles Area). The GCNN model may naturally be the natural choice in this case, as the model size is only \~40 times larger than the MLP one, however GCNN can capture spatio-temporal dependencies of the whole study area, and reports the best results for long-term prediction. The drawback of the GCNN approach resides in data preparation and in the requirement of the road network as input. 

\begin{table}[htbp]
	\centering
	\caption{The training times reported are the average training time per model in seconds. The model size reported is the average model size and is expressed in Kb.}
	\label{tab:pred-stats}
	\begin{tabular}{l|rrr}
		Predictors  & \multicolumn{1}{l}{Training time} & \multicolumn{1}{l}{Size} & \multicolumn{1}{l}{\# of models} \\\hline    
		\rev{SVR} &    \rev{28.8}      &   \rev{8236}    & \rev{$O(Nh)$} \\
		GB   &    \rev{51.9}      &   \rev{4751}    & $O(Nh)$ \\
		AR    &   30.2        &  9     &  $O(N)$ \\
		MLP   & 315.4      & 17    & $O(N)$ \\
		LSTM  & 1892.6    & 800   & $O(N)$  \\
		C-MLP  & 36147.6       & 10    & $O(C)$ \\
		B-MLP & 5352.2       & 38    & $O(1)$ \\
		GCNN  & 18892.4    &   652    &  $O(1)$  \\
	\end{tabular}%
\end{table}%

\rrev{Finally, from an operational point of view, we would like to emphasize that, even though there are wrong models, there is no right model. The selection of the right model will depend on the priority of the practitioner. For instance, if providing the best accuracy is not a hard requirement, a simple contextual average may be the inexpensive go-to solution. In other cases model selection should depend on whether the practitioner has the distributed and streaming infrastructure to maintain, train, and test as many models as number of links (higher than, or equal to thousands), or whether it is better to spend some effort setting up a heavier data transformation step, having a lengthier, but centralized, training process, and maintain and score one unique GCNN model per network.}
	\section{Conclusions}\label{sec:conclusions}

 The implementation of transport and intelligent systems services calls for accurate and fast predictors that are able to scale to a city-level and be reliable for long trips. While short-term traffic predictions have been extensively studied with approaches such as autoregressive models, decision trees, feedforward and recurrent neural networks, in this paper we have looked into longer term predictions for a large urban area of interest. We have provided an evaluation of several literature machine learning and deep learning predictors for link-based prediction in an area with $1098$ links in Los Angeles, California. We have investigated ways to make such predictors scale with clustering, and model design methods. \rrev{The conclusions are three-fold: (i) for short-term prediction and when scalability is not a concern, deep learning techniques do not appear to be particularly useful; (ii) for long-term prediction and/or when scalability is a concern, deep learning techniques may be proven to be particularly useful, as they provide the best accuracy performance for longer prediction time horizons, and as they give the ability to train network models, i.e. have one unique model per area as in the case of our proposed GCNN model; (iii) the recommendations for the practitioner will always depend on the time horizon considered and on the scale of the problem, and when scalability does matter, on the infrastructure available to process data, train and score models.} We should stress that these conclusions are valid for the dataset investigated, and may be less valid across different datasets. In our analysis, the MLP predictors provide a good trade-off between training time, model size, and accuracy. The LTSM predictors, as well as the GCNN predictor we presented, proved to be the most accurate predictors for prediction horizons higher than $2$ hours ahead. Future work will include the adoption of other ensemble methods and different clustering methods to improve accuracy, while keeping overall performance in mind. The integration of LSTM and GCNN, which provided the best prediction results for long-term prediction, should also be investigated, also from a scalability perspective.

	\section{Acknowledgments}\label{sec:acknow}

The authors are extremely grateful to Shoichiro Watanabe, Yohkichi Sasatani, Yasuhisa Gotoh, and Sanehiro Furuichi for their valuable help and feedback throughout this work. 
	
	
	%
	%
	%
	%

	\bibliography{ref}

\end{document}